# End-to-end 3D face reconstruction with deep neural networks

Pengfei Dou, Shishir K. Shah, and Ioannis A. Kakadiaris
Computational Biomedicine Lab
University of Houston
4800 Calhoun Road, Houston, TX 77004
{pdou,sshah,ikakadia}@central.uh.edu

## Abstract

*Monocular 3D facial shape reconstruction from a single 2D facial image has been an active research area due to its wide applications. Inspired by the success of deep neural networks (DNN), we propose a DNN-based approach for End-to-End 3D FAce Reconstruction (UH-E2FAR) from a single 2D image. Different from recent works that reconstruct and refine the 3D face in an iterative manner using both an RGB image and an initial 3D facial shape rendering, our DNN model is end-to-end, and thus the complicated 3D rendering process can be avoided. Moreover, we integrate in the DNN architecture two components, namely a multi-task loss function and a fusion convolutional neural network (CNN) to improve facial expression reconstruction. With the multi-task loss function, 3D face reconstruction is divided into neutral 3D facial shape reconstruction and expressive 3D facial shape reconstruction. The neutral 3D facial shape is class-specific. Therefore, higher layer features are useful. In comparison, the expressive 3D facial shape favors lower or intermediate layer features. With the fusion-CNN, features from different intermediate layers are fused and transformed for predicting the 3D expressive facial shape. Through extensive experiments, we demonstrate the superiority of our end-to-end framework in improving the accuracy of 3D face reconstruction.*

## 1. Introduction

Three-dimensional information, being a strong prior invariant to view perspectives, has been demonstrated beneficial in different computer vision applications [29, 9, 11, 17, 18, 15]. Among these applications, 3D data has been widely employed in face recognition research to address pose, expression, and illumination variations in facial images, resulting in many publications with state-of-the-art performance [11, 17, 15, 34, 8, 10, 5, 6]. In these methods, one crucial step is acquiring the personalized 3D face model which, ideally, can be captured with a 3D camera system. However, the high cost and limited effective sensing range of 3D cameras have constrained their applicability in practice. An alternative approach is reconstructing the 3D facial shape using 2D facial images, which has found wide applications in both the computer vision and the computer graphics communities.

Three-dimensional facial shape reconstruction from 2D image(s) is very challenging by its nature if no prior knowledge is provided. This is mainly due to the large solution space of the problem and the loss of depth information in the image acquisition process. Given prior knowledge about the camera intrinsic parameters or multi-view images of the same subject, a number of methods including multi-view stereo, photometric stereo, or structure-from-motion can be applied to reconstruct the 3D face. However, in most scenarios, camera intrinsic parameters are unknown and usually only a single 2D image is available, making the problem, referred to as monocular 3D facial shape reconstruction (MFSR), even harder.

A common prior employed in solving the monocular 3D facial shape reconstruction problem is the subspace or morphable model [3] that captures shape variations in human face with a set of basis shapes. By using a morphable model, a 3D human face can be parameterized as a vector of weights for the shape basis. As a result, the solution space becomes numerically constrained, thus is solvable by common optimization techniques. To retrieve the optimal model parameters that best reconstruct the 3D facial shape of the input 2D image, Blanz and Vetter [3] proposed to minimize the discrepancy between the input 2D image and the 3D face rendering in an analysis-by-synthesis manner. Though interesting results have been achieved, this method cannot handle complex illumination conditions and suffers from high computation cost. To solve its limitation, Blanz *et al.* [2] proposed to predict the model parameters via linear regression from facial feature point locations. Though efficient, this method abandons most of the useful information in the image and learns very simple regressor functions. As a result, the reconstruction is usually very coarse and sensi-tive to inaccurate facial landmarks.

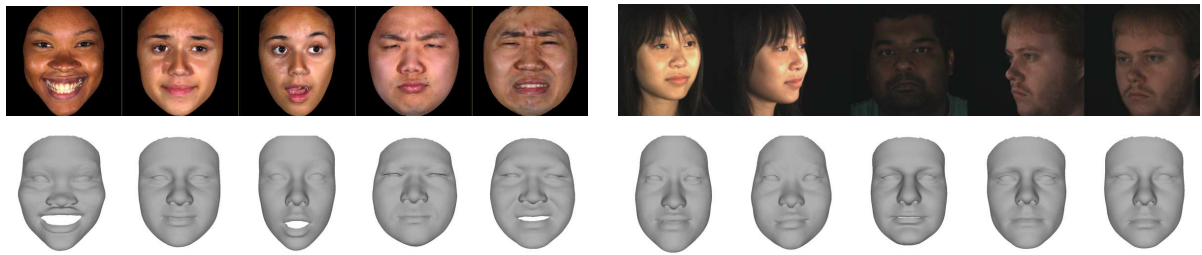

Figure 1: Example of 3D faces reconstructed by our method: (L) Expressive 3D faces reconstructed from facial images with expression and (R) neutral 3D faces reconstructed from images with different facial pose.

Although the past half-decade has witnessed the rapid growth and great success of deep learning in different computer vision research areas, such as object detection and recognition, image segmentation, and image captioning, only limited research in monocular 3D facial shape reconstruction using deep learning exists. With millions of parameters, deep neural networks can be trained to approximate very complex non-linear regressor functions that map a 2D facial image to the optimal morphable model parameters. In this paper, inspired by three recent papers [33, 11, 25], we propose a DNN-based approach: End-to-End 3D FAce Reconstruction (*UH-E2FAR*), that reconstructs the 3D facial shape from a single 2D image. Though sharing the same principal idea, our method differs from [33, 11, 25] in several ways. First, compared with [33, 11, 25], we greatly simplify the framework. Instead of using the iterative scheme employed in [33, 11, 25], our approach is end-to-end and predicts the optimal morphable model parameters with a single forward operation. Instead of using additional data, such as the geometry image employed in [33] or the rendering of an initial 3D face employed in [25], our network takes only an RGB image patch with the facial region-of-interest (ROI) detected as input. With these simplifications in the framework, our approach eases the training of deep neural network and makes it possible to use available 2D facial databases as additional training data to initialize the network. Second, as the 3D facial shape consists of two parts, namely identity and expression, we divide the problem of 3D face reconstruction into two sub-tasks, namely reconstructing the neutral 3D facial shape and reconstructing the facial expression, and incorporate a multi-task learning loss in our approach to train different layers for predicting the identity and the expression parameters separately, which has been demonstrated effective in different applications [30, 23]. To validate our simplifications and modifications, we perform extensive experiments with different neural network architectures and compare their performance with ours.

The rest of the paper is organized as follows. Section 2 reviews related work. Section 3 describes the details of our proposed method. Section 4 describes the implementation details and provides extensive experimental evaluations. Finally, Section 5 concludes the paper with a brief summary and discussion.

## 2. Related work

Blanz and Vetter *et al.* [3] proposed a 3D morphable model (3DMM) for modeling 3D human face from a single or multiple facial images. The model parameters are optimized in an analysis-by-synthesis manner to approximate the input 2D facial image. Though interesting results are achieved, this method suffers from high computational cost and requires manual manipulation to align the mean 3D facial shape to the 2D facial image during initialization. The 3DMM method was extended by Blanz *et al.* [2] to use a sparse set of facial feature points for model parameter estimation. Similarly, Rara *et al.* [24] proposed a regression model between the 2D facial landmarks and the 3DMM parameters and employed principal component regression (PCR) for model parameter estimation. As large facial pose variation might degrade 2D facial landmark detection, Dou *et al.* [7] proposed a dictionary-based representation of 3D facial shape and employed sparse coding to estimate model parameters from facial landmarks. Compared with the PCA-based model, their method achieves better robustness to inaccurate facial landmark detection. Similarly, Zhou *et al.* [32] also employed a dictionary-based model and proposed a convex formulation to estimate model parameters from facial landmarks.

Compared with facial feature points, facial image provides more information useful for reconstructing the 3D face. Wang and Yang [28] proposed to learn mappings from 2D images to corresponding 3D shapes via manifold learning and alignment. Song *et al.* [26] proposed a coupled radial basis function network (C-RBF) method to learn the intrinsic representations and corresponding non-linear mapping functions between paired 2D and 3D training data. Similarly, Liang *et al.* [16] combined RBF network and a coupled dictionary model to reconstruct 3D faces and enhance the facial details for facial expression synthesis. However, these method cannot handle non-frontal facial images. Zhu *et al.* [35] proposed a discriminative approach for 3DMM fitting by using local facial features and a cascade of regressors to estimate and update the 3DMM parameters iteratively. This work was extended by Zhu *et al.* [33] by using deep neural networks to approximate the regression functions. Another work sharing a similar idea is presented in [11]. The results in [33, 11] are inspiring, as they have demonstrated the effectiveness of deep neural networks in approximating the complex mapping function from 2D facial appearance to 3DMM parameters. However, both works focus on 2D face alignment and no experimental evaluation is provided to analyze the performance of their methods in 3D face reconstruction. Richardson *et al.* [25] also proposed using deep neural networks to learn the regression function for estimating 3DMM parameters from 2D facial images. Similar to [33], they use both an RGB image and the 3D rendering of an initial 3D face as input to the network. Different from [33], [25] also inputs the initial 3DMM parameters to establish a feedback loop and force the deep neural networks to iteratively update the 3DMM parameters.

## 3. Method

Similar to [33, 11, 25], we employ a 3D facial shape subspace model and represent the 3D face as the linear combination of a set of shape and blendshape basis:

$$S = \bar{S} + U_d \cdot \alpha_d + U_e \cdot \alpha_e \ , \tag{1}$$

where $S$ is the target 3D face, $\bar{S}$ is the mean facial shape, $U_d$ is the principal components trained on neutral 3D facial scans and $\alpha_d$ is the identity parameter vector, and $U_e$ is the principal components trained on the offset between expressive and neutral 3D facial scans and $\alpha_e$ is the expression parameter vector. Given a 2D facial image, our goal is to predict the optimal identity and expression parameters that minimize the difference between the reconstructed 3D face and the ground truth.

We use two 3D facial shape models, namely the BFM model proposed by [20] and the AFM model proposed by [12]. The BFM model consists of 53,490 vertices and 160,470 triangular faces. It preserves fine facial shape details, thus is very suitable for 3D face modeling and synthesis [33]. The AFM model consists of 7,597 vertices and 14,912 triangular faces. It is a lightweight 3D facial shape model and features a bijective mesh parameterization, thus is very useful for 3D face analysis [12, 27] and 3D-aided 2D facial pose normalization [13, 8].

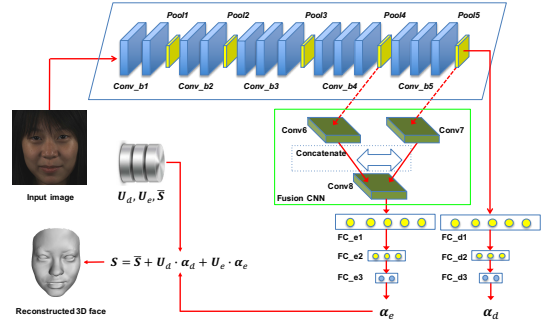

Figure 2: Depiction of the architecture of our deep neural network model for 3D facial shape reconstruction from a single 2D facial image.

### 3.1. Deep neural network architecture

The architecture of our deep neural network is illustrated in Fig. 2. It is based on the *VGG-Face* model [19] that consists of 13 convolutional layers and 5 pooling layers. Specifically, we add two key components, a sub convolutional neural network (**fusion-CNN**) that fuses features from intermediate layers of *VGG-Face* for regressing the expression parameters and a multi-task learning loss function for both the identity parameters prediction and the expression parameters prediction. With both components, we can train three types of neural layers in a single DNN architecture. The first type of neural layers includes those below the fourth pooling layer (Pool4), which learn generic features corresponding to low-level facial structures, such as edges and corners. These layers are shared by the two tasks. The second type of neural layers includes the three convolutional layers in the *fusion-CNN* and the following fully connected layers. These layers are forced to learn expression-specific features. The third type of neural layers includes those above the fourth pooling layer (Pool4) which learn class-specific features that are more suitable for predicting the identity parameters.

The input to the network is an RGB image cropped and scaled to $180 \times 180$ pixels. To fuse the intermediate features from layer Pool4 and layer Pool5, we set the kernel size and stride of layer Conv6 and layer Conv7 to be $\{5 \times 5, 2\}$ and $\{1 \times 1, 1\}$, respectively. After concatenating the features from Conv6 and Conv7, we add another $1 \times 1$ kernel convolutional layer Conv8 to reduce the feature dimension. The details of all layers (except for those in the backbone) are summarized in Table 1.

### 3.2. The end-to-end training

The input to our deep neural network is an 2D image with the facial ROI localized by a face detector. In this pa-per, we use the Dlib SDK[1] for face detection. We first enlarge the detected face bounding box by a factor of 0.25 of its original size and then extend the shorter edge to crop a square image patch of the face ROI, which is scaled to be $180 \times 180$. The output of the deep neural network consists of the identity parameter vector and the expression parameter vector. They are used to reconstruct the 3D facial shape corresponding to the input 2D image using Eq. 1.

| Layer | Conv6 | Conv7 | Conv8 | FC_e1 | FC_e2 | FC_e3 | FC_d1 | FC_d2 | FC_d3 |
|---|---|---|---|---|---|---|---|---|---|
| Input Size | $512 \times 12 \times 12$ | $512 \times 6 \times 6$ | $1024 \times 6 \times 6$ | $512 \times 6 \times 6$ | 4,096 | 1,024 | $512 \times 6 \times 6$ | 4,096 | 1,024 |
| Output Size | $512 \times 6 \times 6$ | $512 \times 6 \times 6$ | $512 \times 6 \times 6$ | 4,096 | 1,024 | 29 | 4,096 | 1,024 | 199 |
| Stride, Pad | 2, 2 | 1, 0 | 1, 0 | N/A | N/A | N/A | N/A | N/A | N/A |
| Filter Size | $5 \times 5$ | $1 \times 1$ | $1 \times 1$ | N/A | N/A | N/A | N/A | N/A | N/A |

Table 1: Specifications of different layers in our deep neural network architecture.

**Training data:** We propose using both real 2D images and synthetic 2D images to train the deep neural network. Real 2D images are used to initialize the deep neural network and synthetic 2D images are used for fine-tuning. We follow a similar procedure as Richardson *et al.* [25] to generate synthetic facial images for training our deep neural network. For the BFM 3D facial shape model, we use the shape basis provided by [20] and the blendshape basis provided by [33, 4]. For the AFM 3D facial shape model, we select 203 neutral 3D facial scans from the FRGC2 [21] and the BU-3DFE [31] databases and register them using the fitting algorithm proposed by [12] to build the shape basis. For simplicity, we do not build the blendshape basis for the AFM model. In total, we create 10,000 random neutral 3D faces for both the BFM and the AFM 3D facial shape models, corresponding to 10,000 identities. For each 3D face we synthesize 25 images with different facial pose, illumination, and facial expression. More details are presented in Sec. 4.1.

**Cost functions:** We choose the training cost as the difference between the predicted 3D face and the ground truth. To measure this difference, we employ the sum of squared error over all 3D vertices:

$$E_c = \|\boldsymbol{U}_c \cdot \hat{\boldsymbol{\alpha}}_c - \boldsymbol{U}_c \cdot \boldsymbol{\alpha}_c\|_2^2 \ , \quad (2)$$

where $c \in \{e, d\}$, $\hat{\boldsymbol{\alpha}}_c$ denotes the predicted parameter vector, and $\boldsymbol{\alpha}_c$ denotes the ground truth.

The total loss is computed as the weighted sum of both losses:

$$E = \lambda_d E_d + \lambda_e E_e \ , \quad (3)$$

where $\lambda_d$ and $\lambda_e$ are weights for the two separate losses.

---
[1] http://dlib.net/

### 3.3. Discussion

Compared with [33, 11, 25], one major difference of our deep neural network is that it is end-to-end and takes only a single RGB image as input. As a result, the training and deployment of our DNN model is greatly simplified. During training, we do not need to perform iterative data simulation as [33, 25], which is non-trivial. During deployment, we do not need to perform 3D rendering of the initial or intermediate 3D facial shape, which is costly and might be restricted in certain scenarios. Another benefit of using a single RGB image as input is that we can use available 2D face databases to initialize our DNN model, which helps improve the robustness of our approach to facial pose and the complex illumination conditions. Another difference is that we employ a multi-task learning loss and a fusion-CNN for fusing intermediate features. As a result, we are able to train different layers for predicting the identity and the expression parameters separately. Intermediate features carry important discriminative information useful for expression parameter vector prediction. In comparison, high-level features are class-specific and robust to facial expression variations, thus are beneficial for predicting the identity parameter vector.

## 4. Experiments

In this section, we evaluate our *UH-E2FAR* algorithm for 3D face reconstruction from a single image. We compare it with several state-of-the-art algorithms, namely *RSNIEF* [25], *RSN*, and *UH-2FCSL* [7]. The *RSN* algorithm is a modification of *RSNIEF* by removing the feedback connection and using only 2D images without 3D synthetic renderings. We also compare our method with the *UH-E2FARMod* algorithm, a modification of our *UH-E2FAR* algorithm by removing the fusion convolutional neural networks (**fusion-CNN**) to demonstrate the advantage of our algorithm in reconstructing expressive 3D faces.

### 4.1. Synthetic data generation

Due to the unavailability of large scale 3D-2D face databases, we follow a similar procedure as [25] to create synthetic training data. As mentioned in Sec. 3.2, we first create the 10,000 neutral 3D faces and their corresponding facial textures using random parameters. Then, we proceed to generate various facial expressions by changing the expression parameters. We have observed that using a random expression parameter will generate a lot of non-plausible 3D facial shapes. As a result, we collect a very large set of expression parameters estimated on multiple 2D face databases by [33] and sample them randomly. In this way, we ensure that the facial expression generated by the sampled expression parameter will be plausible.

To generate plausible synthetic images, it is crucial to control the camera parameters and illumination properly during 3D rendering. We use a perspective camera model and set the camera field-of-view randomly to be within the range of $[15°, 35°]$. Accordingly, the distance between the camera and the object is set to be within 1,900 *mm* and 500 *mm*. We use the Phong Reflectance model [22] for illumination synthesis. For the shininess parameter, instead of using a constant for all 3D faces, we also collect a large set of shininess parameters estimated on two 2D face databases by [20] and sample them randomly. For the ambient, diffuse, and specular parameters, we use random values within the ranges $[0.2, 0.4]$, $[0.6, 0.8]$, and $[0.1, 0.2]$. The facial poses of the synthetic images are randomly generated. The yaw, pitch, and roll rotations are uniformly distributed within the ranges $[-90°, 90°]$, $[-30°, 30°]$, and $[-30°, 30°]$. The background of the synthetic images is also randomly generated. Examples of the generated synthetic images corresponding to the BFM and the AFM 3D facial shape models are depicted in Fig. 3.

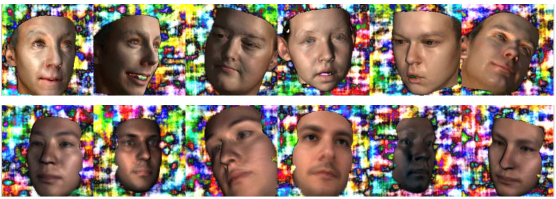

Figure 3: Example of generated synthetic facial images corresponding to the two 3D facial shape models employed: (T) The BFM facial shape model and (B) the AFM facial shape model.

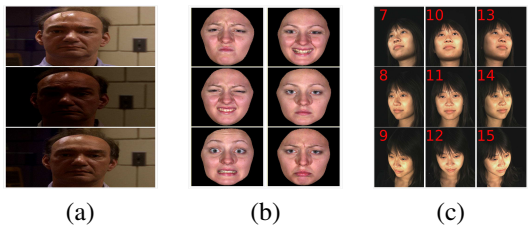

Figure 4: Example of 2D images from the three public databases used in experiments: (a) The FRGC2 database, (b) the BU-3DFE database, and (c) the UHDB31 database.

### 4.2. Evaluation databases and metrics

In addition to the synthetic data, we use three publicly available 3D face databases in our experiments, namely the FRGC2 database [21], the BU-3DFE database [31], and the UHDB31 database [29]. For the FRGC2 database, we use the validation partition that consists of 4,007 pairs of 2D and 3D data of 466 subjects. The 2D facial images are captured under different illumination conditions. For the BU-3DFE database, we use all the 2,500 pairs of 2D and 3D data of 100 subjects. The 2D and 3D data are captured while the subjects are performing different types of facial expressions. For the UHDB31 database, we use a subset of 2,079 2D facial images with corresponding 3D facial scans. These data are captured under three illumination conditions with nine facial poses. We show several examples of the 2D data from these three databases in Fig. 4.

In the first experiment, these three databases are all used to evaluate and compare the performance of our method with state-of-the-art using the BFM 3D facial shape model. The FRGC2 database is used to evaluate the performance of different methods under varying illumination conditions. The BU-3DFE database is used to evaluate their performance with varying facial expressions. The UHDB31 database is used to evaluate their performance with different facial poses. In the second experiment, the FRGC2 and the BU-3DFE databases are used to build the shape basis of the AFM 3D facial shape model, while the UHDB31 database is used for evaluation.

To compare the performance of different methods, we used the root mean squared error between the reconstructed 3D face and the ground truth after rigid alignment and registration using the iterative closest point (ICP) algorithm [1] to measure the accuracy of 3D face reconstruction.

### 4.3. Implementation details

We use the Caffe deep learning framework[2] to train the four DNN models. The pre-trained *VGG-Face* model is used as initialization for *UH-E2FAR* and *UH-E2FARMod*, which are then fine-tuned on the synthetic database for 3D face reconstruction. The multi-task loss weights of *UH-E2FAR* are empirically set to $\lambda_d = 1$ and $\lambda_e = 5$. The Adam solver [14] is employed with the mini-batch size and the initial learning rate set to 32 and 0.0001, respectively. We first fine-tune only the fully connected layers and the fusion-CNN for 40,000 iterations. Then, we continue to fine-tune the full deep neural network. For *RSNIEF* and *RSN*, we train them on the synthetic database for 3D face reconstruction from scratch. The Adam solver [14] is employed with the initial learning rate set to 0.001. The learning rate is decreased by a factor of 0.5 every 40,000 iterations. We run the training for a total of 120,000 iterations.

---

[2]http://caffe.berkeleyvision.org/

|  | UH-E2FAR | RSNIEF | RSN | UH-2FCSL |
|---|---|---|---|---|
| **UHDB31** | **2.73±0.71** | 3.51±0.84 | 3.65±0.91 | 3.37±0.76 |
| **FRGC2** | **3.71±3.05** | 3.91±2.51 | 4.50±3.09 | 3.81±2.30 |
| **BU-3DFE** | 4.52±1.11 | **4.00±1.07** | 4.23±1.09 | N/A |

Table 2: Quantitative comparison on the UHDB31, the FRGC2, and the BU-3DFE databases: Mean and standard deviation of RMSE (*mm*).

were used to reconstruct the 3D faces. The cumulative distribution of RMSE is depicted in Fig. 5. The quantitative results of mean and standard deviation of RMSE are illustrated in Table 2. The spatial reconstruction error distributions of the four methods over the facial region are depicted in Fig. 6(T) and the comparisons between our method *UH-E2FAR* and the other three methods are depicted in Fig. 6(B). It is clear that our method offers the best performance.

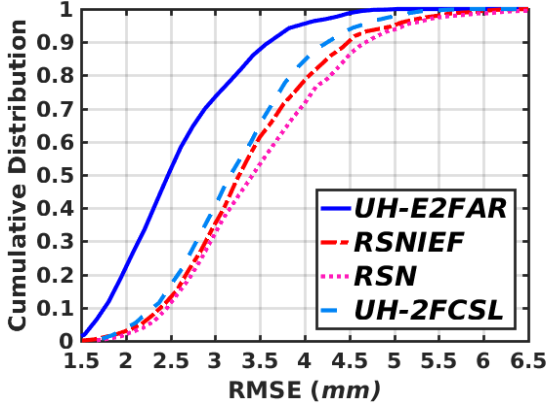

Figure 5: Cumulative distribution of 3D facial shape reconstruction RMSE on the UHDB31 database.

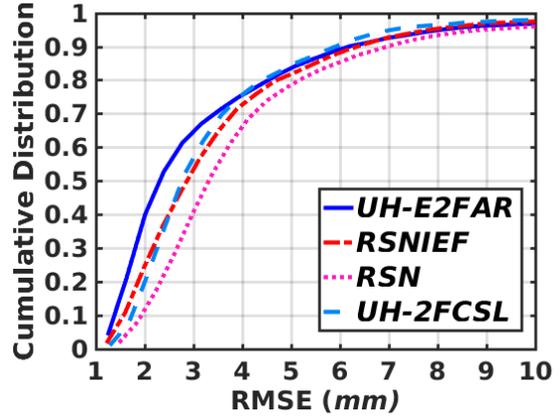

Figure 7: Cumulative distribution of 3D facial shape reconstruction RMSE on the FRGC2 database.

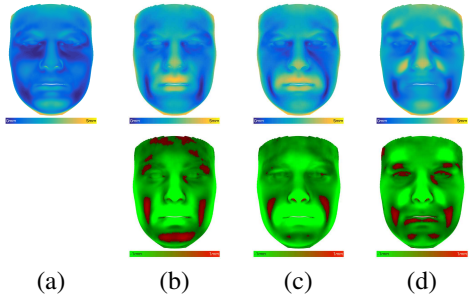

(a)　　(b)　　(c)　　(d)

Figure 6: Reconstruction error heatmaps of different methods on the UHDB31 database: (a) *UH-E2FAR*, (b) *RSNIEF*, (c) *RSN*, and (d) *UH-2FCSL*. The top row illustrates the spatial distribution of RMSE on the face and the bottom row illustrates the differences in RMSE between our approach *UH-E2FAR* and other approaches (green indicating our method has smaller RMSE, red indicating our method has larger RMSE, and color intensity indicating the magnitude of the difference).

### 4.4. Experimental Results

In the first experiment, we evaluated the performance of our method on the multi-view facial images in the UHDB31 database. In total, 1,638 out of the selected 2,079 2D facial images with successful face and facial landmark detection

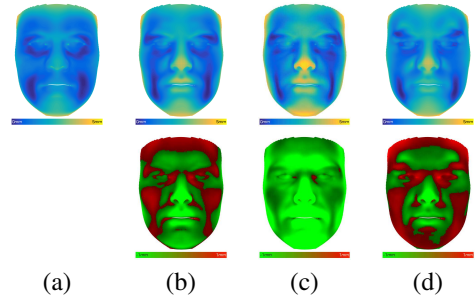

(a)　　(b)　　(c)　　(d)

Figure 8: Reconstruction error heatmaps of different methods on the FRGC2 database: (a) *UH-E2FAR*, (b) *RSNIEF*, (c) *RSN*, and (d) *UH-2FCSL*. The top row illustrates the spatial distribution of RMSE on the face and the bottom row illustrates the differences in RMSE between our approach *UH-E2FAR* and other approaches (green indicating our method has smaller RMSE, red indicating our method has larger RMSE, and color intensity indicating the magnitude of the difference).

Similarly, on the FRGC2 database, 3,999 out of the 4,007 facial images with successful face and facial landmark detection are used to evaluate the 3D face reconstruction accuracy. The cumulative distribution of RMSE is depicted in Fig. 7. The quantitative results of mean and standard deviation of RMSE are illustrated in Table 2. Compared with *RSNIEF*, *RSN*, and *UH-2FCSL*, our algorithm achieves considerable improvement in reconstruction accuracy. The spatial reconstruction error distributions of the four methods over the facial region are illustrated in Fig. 8(T) and the comparisons between our method *UH-E2FAR* and the other three methods are depicted in Fig. 8(B). Compared with *RSNIEF* and *UH-2FCSL*, our method demonstrates better performance at the key facial regions including the mouth, nose, and eyes.

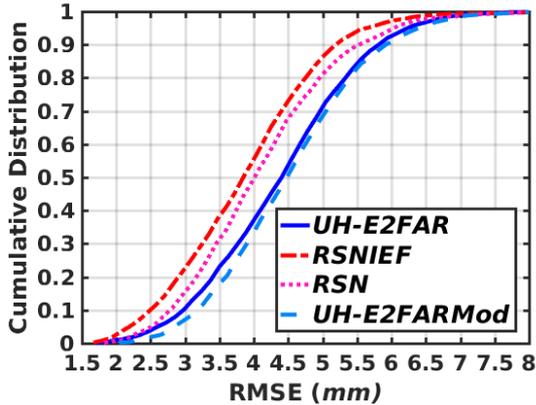

Figure 9: Cumulative distribution of 3D facial shape reconstruction RMSE on the BU-3DFE database.

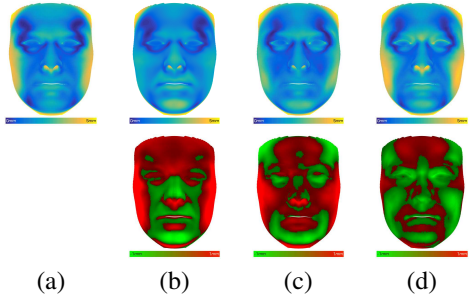

Figure 10: Reconstruction error heatmaps of different methods on the BU-3DFE database: (a) *UH-E2FAR*, (b) *RSNIEF*, (c) *RSN*, and (d) *UH-E2FARMod*. The top row illustrates the spatial distribution of RMSE on the face and the bottom row illustrates the differences in RMSE between our approach *UH-E2FAR* and other approaches (green indicating our method has smaller RMSE, red indicating our method has larger RMSE, and color intensity indicating the magnitude of the difference).

We evaluate and compare our method with *RSNIEF* and *RSN* in reconstructing expressive 3D face on the BU-3DFE database. As *UH-2FCSL* and *DRSN* are not capable of reconstructing expressive 3D face, we exclude these two methods from this experiment. The cumulative distribution of RMSE is depicted in Fig. 9. The quantitative results of mean and standard deviation of RMSE are illustrated in Table 2. Compared with *RSNIEF* and *RSN*, our method exhibits larger RMSE. From the spatial reconstruction error distributions over the facial region, as depicted in Fig. 10(T), we observe that the majority of RMSE error of *UH-E2FAR* is distributed in the outer facial region. In the inner facial region, our method exhibits much lower RMSE compared with *RSNIEF* and comparable RMSE when compared with *RSN*. We also compare our method with *UH-E2FARMod* to demonstrate the benefit of the fusion-CNN we proposed in reconstructing expressive 3D face. Compared with *UH-E2FARMod*, *UH-E2FAR* exhibits better performance at the key facial regions including the mouth, nose, and eyes.

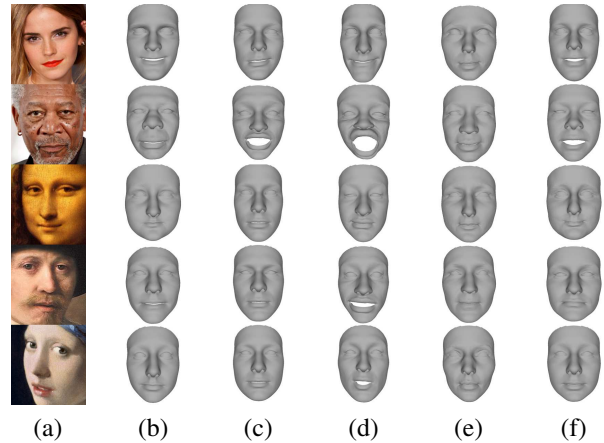

Figure 11: Example of the neutral 3D faces reconstructed by different methods: (a) The input 2D image. (b) *UH-E2FAR*, (c) *RSNIEF*, (d) *RSN*, (e) *UH-2FCSL*, and (f) *UH-E2FARMod*.

The neutral 3D faces reconstructed by different methods from facial images captured in the wild are depicted in Fig. 11. Compared with *RSNIEF*, *RSN*, and UH-E2FARMod, *UH-E2FAR* is more robust. The expressive 3D faces reconstructed by different methods are illustrated in Fig. 12. Note that the performance of our method is very stable and the reconstructed expression is more plausible than *RSNIEF* and *RSN*, and more accurate than *UH-E2FARMod*.

In the second experiment, we integrate our approach into the 2D face recognition pipeline proposed by Kakadiaris *et al.* [13] and compare with *UH-2FCSL* for 3D-aided face recognition. We use the AFM 3D facial shape model and train our *UH-E2FAR* model with 250,000 synthetic images. We use the frontal facial image of each subject in the UHDB31 database as gallery and use the other eight non-frontal images as probe. To emphasize the influence of 3D face reconstruction on face recognition performance, we use manually annotated feature points on 2D facial images during 3D2D pose estimation and employ a simple facial representation based on image gradient computed on the pose-normalized facial texture. Besides *UH-E2FAR* and *UH-2FCSL*, we also use the ground truth 3D facial data as a baseline to highlight the performance of our method. The cumulative match characteristic (CMC) curve of face identification accuracy is depicted in Fig. 13. Compared with *UH-2FCSL*, our DNN-based approach *UH-E2FAR* increases the face identification accuracy considerably. From the fine-grained results of rank-1 face identification rates on different facial poses, as depicted in Fig. 14, we observe that in some facial poses the rank-1 identification rates obtained with our reconstructed 3D faces are very close to those obtained using ground truth 3D data. It indicates the superior performance of our approach.

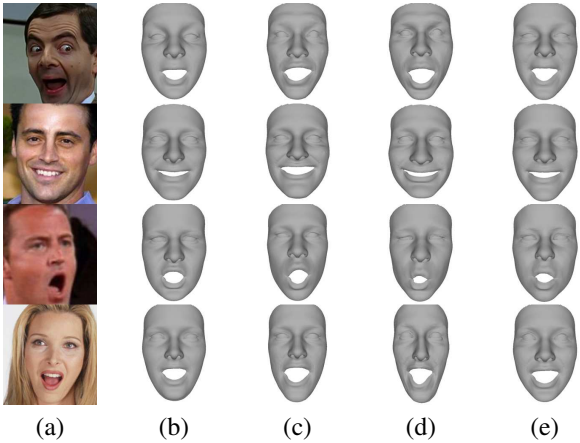

(a) (b) (c) (d) (e)

Figure 12: Example of the expressive 3D faces reconstructed by different methods: (a) The input 2D image. (b) *UH-E2FAR*, (c) *RSNIEF*, (d) *RSN*, and (e) *UH-E2FARMod*.

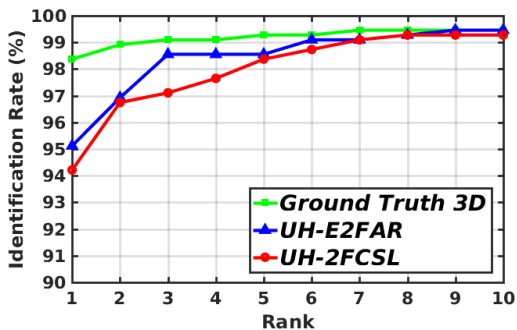

Figure 13: The cumulative matching characteristic curve of face identification accuracy on the UHDB31 database.

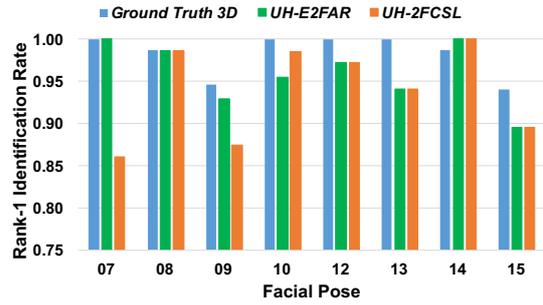

Figure 14: The Rank-1 identification rates on different facial poses.

## 5. Conclusions

In this paper, we propose *UH-E2FAR*, an end-to-end 3D face reconstruction method based on deep neural networks. Compared with previous work, our method brings significant simplification to the framework by replacing the iterative model parameter updating scheme with an end-to-end inferring scheme and removing the dependency on 3D shape rendering or initial model parameters as additional DNN input. We also introduce two key components to our framework, namely a fusion-CNN and a multi-task learning loss. With both components, we divide 3D face reconstruction into two sub-tasks, namely neutral 3D facial shape reconstruction and expressive 3D facial shape reconstruction, and train different types of neural layers in a single DNN model for these two specific tasks. With extensive experiments, we demonstrate that the simplification of the framework does not compromise the 3D face reconstruction performance. Instead, it is possible to initialize our DNN model using real facial images from available 2D face databases, which helps improve the robustness of our method to facial pose and complex illumination. As a result, our method outperforms state-of-the-art approaches [25, 7] with significant improvement in reconstruction accuracy and robustness.

## Acknowledgment

This material is based upon work supported by the U.S. Department of Homeland Security under Grant Award Number 2015-ST-061-BSH001. This grant is awarded to the Borders, Trade, and Immigration (BTI) Institute: A DHS Center of Excellence led by the University of Houston, and includes support for the project "Image and Video Person Identification in an Operational Environment: Phase I" awarded to the University of Houston. The views and conclusions contained in this document are those of the authors and should not be interpreted as necessarily representing the official policies, either expressed or implied, of the U.S. Department of Homeland Security.